\documentclass[conference]{IEEEtran}
\IEEEoverridecommandlockouts

\usepackage{cite}
\usepackage{amsmath,amssymb,amsfonts}
\usepackage{algorithmic}
\usepackage{graphicx}
\usepackage{textcomp}
\usepackage{xcolor}

\usepackage[T1]{fontenc}
\usepackage{graphicx}
\usepackage{hyperref}
\usepackage{fvextra}
\usepackage{booktabs}
\usepackage{siunitx}
\usepackage{multirow}

\usepackage{booktabs} 

\usepackage{mdframed}
\usepackage{xcolor}
\usepackage{listings}

\newmdenv[
  linecolor=black,
  linewidth=1pt,
  topline=true,
  bottomline=true,
  leftline=true,
  rightline=true,
  innerleftmargin=10pt,
  innerrightmargin=10pt,
  innertopmargin=8pt,
  innerbottommargin=8pt,
  backgroundcolor=gray!8,
]{signaturebox}

\def\BibTeX{{\rm B\kern-.05em{\sc i\kern-.025em b}\kern-.08em
    T\kern-.1667em\lower.7ex\hbox{E}\kern-.125emX}}
\begin{document}

\title{Large Language Models for Programming: Actually Fixing or Reimplementing Incorrect Code?}

\author{
\IEEEauthorblockN{Alexandru Stefan Stoica}
\IEEEauthorblockA{\textit{University Politehnica of Bucharest}\\
Bucharest, Romania \\
astoica0909@stud.acs.upb.ro}
\and
\IEEEauthorblockN{Traian Rebedea}
\IEEEauthorblockA{\textit{University Politehnica of Bucharest}\\
Bucharest, Romania \\
traian.rebedea@upb.ro}
\and
\IEEEauthorblockN{Marian Cristian Mihaescu}
\IEEEauthorblockA{\textit{University of Craiova}\\
Craiova, Romania \\
cristian.mihaescu@edu.ucv.ro}
}

\maketitle

\begin{abstract}
Recent studies have shown that Large Language Models can effectively solve problems and fix bugs in diverse programming environments, including competitive programming. Existing approaches primarily evaluate LLM performance in problem solving or bug fixing independently, but do not explore the relationship between these two capabilities. This work focuses on determining how much the LLM deviates from a buggy solution to fix the bug compared to a human-written patch, and if there is a bias towards generating entirely new solutions. We construct a dataset with all the submissions ($\sim$ 3000) from a couple of users from Codeforces, and we match each buggy submission with its corresponding human fix. By using the similarity between the buggy solution and the human fix as a baseline, we evaluate the quality of LLM-generated bug fixes on 3 OpenAI GPT models (gpt-5-nano, gpt-5-mini, gpt-5.1). We check if the generated solutions solve the problem by using the Codeforces-R1 dataset, an openly available dataset that has tests generated with the DeepSeek-R1 model. Our findings suggest that LLMs tend to modify more lines than necessary compared to human fixes and, in some cases, generate entirely new solutions. We also observe that LLMs solve more problems correctly when allowed to generate solutions from scratch rather than patch buggy submissions, even when those submissions are close to the human patch. This has important implications for the design of AI-assisted programming tools, particularly in supporting user debugging processes and promoting incremental problem-solving strategies rather than solution replacement. 
\end{abstract}


\section{Introduction}
Competitive programming is a mind sport in which participants try to come up with solutions to given problems (algorithmic in nature) that must satisfy some constraints. The submission to a given problem is evaluated through multiple test cases that capture various edge cases and inform the user if a constraint is violated. For a given problem to be considered Accepted, one has to pass all the test cases. Besides being a sport, it also represents a method in which a user can learn more about algorithmic design by receiving feedback on their submission by following an iterative debugging process. 


The rapid advancement of Large Language Models (LLMs) has driven significant interest in their application to Automatic Program Repair (APR). Prior work has demonstrated that LLMs can both generate solutions to algorithmic problems~\cite{leblond2023alphacode2} and patch buggy code~\cite{10.1145/3650212.3680323} across a variety of settings. However, existing approaches largely evaluate these two capabilities in isolation, leaving open an important question: when an LLM is given a buggy submission and tasked with fixing it, does it behave like a human debugger — making targeted, minimal edits — or does it tend to discard the original solution and generate a new one from scratch?

This distinction matters beyond mere academic curiosity. In practice, AI-assisted programming tools are increasingly integrated into iterative development workflows, where preserving the user's intent and code structure is often as important as producing a correct solution. A tool that systematically replaces rather than repairs code undermines the debugging process and offers little support for incremental problem solving.

In this paper, we introduce a dataset that contains all the submissions from 7 Codeforces~\cite{codeforces} users ($\sim$ 3000 submissions), which we use to investigate how various strong closed-source LLMs from the OpenAI family (gpt-5-nano, gpt-5-mini, and gpt-5.1) repair their buggy submissions relative to their own fix. In other words, we identify how well the LLM preserves the original structure of the code in the buggy submission when generating a fix, compared to a human patch. Using 3 methods of prompting: (1) a simple bug fix prompt; (2) a prompt that just generate a solution from scratch; (3) an optimized prompt via Genetic-Pareto (GEPA) \cite{agrawal2025gepareflectivepromptevolution}, a prompt optimization technique, we compare the similarity between the buggy problem and the generated fix to the similarity between the buggy problem and the human fix. We also use Codeforces-R1 \cite{penedo2025codeforces} dataset to evaluate the solutions for each problem using coding tests generated by DeepSeek-R1 \cite{Guo_2025} to determine if the code is correct. With this in mind, we investigate the following research questions:
\begin{itemize}
    \item[]
    \begin{itemize}
     \item \textbf{RQ1:} Given a buggy submission, how well does the LLM preserve the original user's structure when fixing the bug?
     \item \textbf{RQ2}: Are language models better at generating code that solves a problem than at fixing bugs?
     \item \textbf{RQ3}: Is there a relationship between how much the AI preserves the user's original code structure and its ability to correctly fix the bug?
    \end{itemize}
\end{itemize}

In summary, this paper makes the following contributions: (1) We introduce a dataset of ~3,000 submissions from 7 Codeforces users, capturing complete submission histories including buggy and accepted solutions. (2) We systematically evaluate three prompting strategies — bug fix, scratch generation, and GEPA-optimized — across three OpenAI models on their ability to repair buggy code. (3) We propose and apply a structural similarity analysis to measure how faithfully LLM-generated fixes preserve the original code, comparing it against human patches. (4) We provide empirical answers to the interplay between structural preservation and correctness, with implications for AI-assisted debugging tools in educational and professional contexts.


\section{Related Work}
\subsection{Evaluation of LLMs in Solving Competitive Programming Problems}
Zheng et al.~\cite{zheng2025livecodebenchproolympiadmedalists} have observed that the current models excel in cases of implementation and derivation of competitive programming solutions, while they perform significantly worse on problems that require heavy observation. At the same time, LLMs alone struggle to solve hard problems or unseen problems, but human-LLM collaboration significantly improves task performance~\cite{yang2025elaborationcomprehensivebenchmarkhumanllm}. The same study finds that this is also true for bug identification. A problem identified by both papers is that data contamination is a common problem in competitive programming evaluation of LLMs. 

\subsection{Evaluation of LLMs in Fixing Bugs in Competitive Programming Problems}
Some recent works explored the capabilities of language models to fix incorrect competitive programming solutions. DebugBench~\cite{tian2024debugbench} is a benchmark with different types of bugs that was constructed via bug implantation with various LLMs on submissions from LeetCode. After a rigorous human evaluation, multiple LLMs are evaluated against the benchmark and the results suggest that LLMs tend to perform better on syntax and reference errors compared to logical errors. Mavalankar et al.~\cite{mavalankar2025aupair} introduce a golden set of example buggy / fix code pairs that is then used to enhance the performance of LLMs in bug fixing.  
In \cite{haque2023fixevalexecutionbasedevaluationprogram} two Language Models (PLBart \cite{ahmad-etal-2021-unified} and CodeT5 \cite{wang2021codet5}) were evaluated in the context of program repair on n-gram based metrics and execution based metrics based on (bug fix, human fix) pairs, and it was found that the n-gram based metrics are suboptimal in comparison with the execution based metrics.
A bug fix should be as localized as possible, and it should not change the structure of the code unless the user explicitly wants this. Most existing works focus on the capacity of LLMs to solve bugs, but not on how they are solved. To fill this gap, our work explores the differences between the human fixes and the generated fixes as well in terms of modified lines and correctness. To this aspect, the only similar work is from Dai et al.~\cite{dai2025less} that has also explored fixing bugs in competitive programming with minimal modifications by defining the consistency of a fix as the ratio between preserved lines and the number of lines in the fixed code. We further expand this metric by also penalizing the lines that are deleted from the initial code. Even though there may be cases where these lines are not actually used anywhere, deleting them is a deviation from what the user asked.


\section{Method}
\subsection {Dataset Creation}
Detecting if a LLM is generating a solution from scratch for a buggy submission instead of actually fixing the bug requires the following: (1) problem specification; (2) a suite of tests on which a generated solution can be verified that is correct; (3) a buggy submission; (4) a human-authored corrected submission that fixes the errors in the buggy submission. The similarity between the buggy submission and the correct one can be used as a threshold $\gamma$ that identifies if the generated solution deviates too much from the buggy submission or makes fewer modifications than the human to fix the bug.

For (1) and (2), we used Codeforces-R1, a dataset with over 10k unique problems extracted from Codeforces along with public test cases and test cases generated with Deepseek-R1. The team behind Codeforces-R1 offers a dataset for submissions (over 12M) as well, but the distributions of verdicts is highly unbalanced, 99\% of the submissions being submissions that have Accepted as a verdict. The rest of submissions are buggy submissions that are sparse and cannot be connected with the correct Accepted submissions, so this dataset is not suitable for (3) and (4).

To gather data for paired buggy (3) and human-authored correct (4)  submissions, we asked various Competitive programming users to provide all the submissions that they have on Codeforces. All the users, started competing on Codeforces when they were students at BSc level (aged 19) and in some cases continuing to do so at MSc level or after. The submissions span from 2015 to 2025. In this way, for a given problem $p$ and a user $u$, we can extract all the submissions and order them in descending order by date. We consider a solution incorrect if it does not pass all the test cases for that specific problem, or if it does not compile. Ordering by date/time is very important in this case, because it offers the possibility to have different snapshots of a given solution from a user, based on submissions. Let us assume that for a given problem $P$, we have $n$ submissions, $s_i$, from the same user $u$, where only the last submission $s_n$ is correct, and the other submissions are wrong. Starting from $s_1$, one needs to add or delete a couple of lines to get to $s_2$. The process is repeated from $s_2$ to $s_3$ and so on till $s_{n-1}$ and $s_n$. In theory, $s_{n-1}$ submission should have the minimal number of lines that need to be added or deleted to reach an Accepted solution, while $s_1$ should have the maximal number of lines that need to be added or deleted. Having these historical submissions, we could estimate a threshold ($\gamma$) that we can use to determine if a solution is generated from scratch by the LLM or it is actually doing a set of minimal modifications to reach a correct solution -- thus being a fix. From now on, we refer to $s_n$ as the anchor solution and $s_i, i \in {1,2..,n-1}$, as being the buggy solutions that are related to $s_n$.

In practice, the following scenarios usually happen: (1) An user could have more than one Accepted solution to a given problem (e.g. trying a different approach); (2) The buggy solutions could be very similar to each other in relation to a given anchor (e.g. there is a small error which the user is not figuring out); (3) The buggy solutions could be very different from the anchor solution. The user has multiple buggy submissions, and then it figures out that the whole approach is wrong and reimplements everything from scratch. To solve (1), we order in descending order and determine the anchor in a greedy manner. Iterating over the submissions, whenever the verdict of a given submission $s_i$ is Accepted, we create a new anchor $a_i$ and we attach to it the following submissions $s_{i+1}, s_{i+2}...$  till we encounter another Accepted solution. Point (2) is not a problem but a useful property. These submissions could be considered as small augmentations of a buggy solution that check the consistency of LLMs' output. For case (3), we choose to retain only the buggy submissions that have a similarity of at least 60 \% with the correct submission. Anything less than 60\% increases the chance of having a re-implementation, where the anchor transforms from a bug fixing solution to a new solution altogether. Comparing the LLM's generated solution to an anchor like this would change the focus from comparing the code in terms of bug fixing modifications to modifications that reach an alternative solution. We consider that a threshold of 60 \% is high enough to avoid cases of re-implementations in anchors but low enough to capture cases in which a LLM genuinely comes up with a bug fixing that is closer to the original code compared to the anchor. (e.g. the user adds or removes a lot of unnecessary code to fix the bug while a LLM modifies only a couple of lines to fix the bug). 

\begin{figure}[h]
    \centering
    \tiny
    \includegraphics[width=0.5\textwidth]{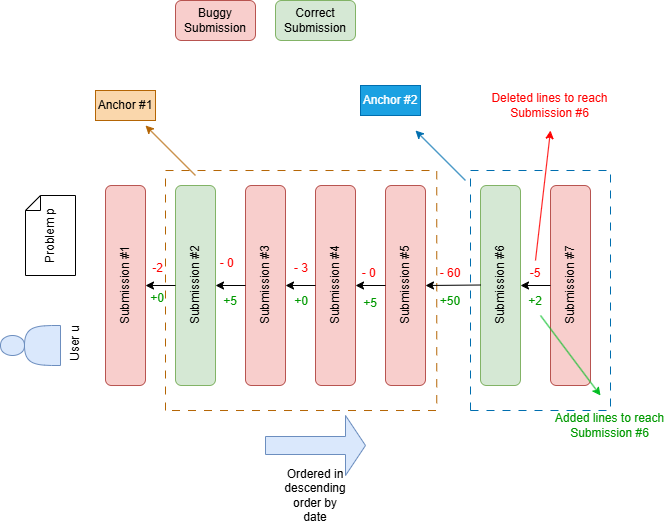} 
    \caption{For a given user $u$ and problem $p$, we order the submissions in descending order. There are two anchors that are related to 3, respectively 1 buggy submission. }
    \label{fig:CalLLMsFixSubmissions_Diagram_submissions}
\end{figure}

Figure \ref{fig:CalLLMsFixSubmissions_Diagram_submissions} depicts the process described in this section for a given problem, where multiple anchors and buggy submissions can be seen. 

\subsubsection {Dataset}
Based on the extracted submissions, we created a dataset with the following features:
\begin{itemize}
    \item[]
    \begin{itemize}
     \item \textit{submissions\_id} - The id of the submission.
     \item \textit{problem\_id} - The id of the problem for the given submission.
     \item \textit{creation\_time\_seconds} - Timestamp when the submission was created.
     \item \textit{programming\_language} - The version of the compiler used to execute the submission.
     \item \textit{verdict} - The verdict of the submission on Codeforces (Accepted, Runtime Error, Memory Limit Exceeded, Time Limit Exceeded, Wrong Answer, Compilation Error, Challenged, Skipped and Rejected).
     \item \textit{source\_code} - The source code of the submission.
     \item \textit{accepted\_anchor} - The submission id to which the given submission is connected to or -1 (the submission is an anchor or it is a buggy submission which has no anchor).
     \item \textit{is\_problem\_usable} - The problem id can be found or not in Codeforces-R1 dataset. 
     \item \textit{passes\_r1\_tests} - The submission passes or not all the test cases.
    \end{itemize}
\end{itemize}

We acknowledge that Codeforces problems may be \textit{contaminated} as they are usually used for training LLMs, but we are less interested in the capacity of LLMs to solve novel problems. Our main focus is on bug fixing with minimal modifications and the relationship between bug fixing and problem solving.

Codeforces evaluates each submission with a variety of tests that capture all the edge cases, but the test cases themselves are partially available (they are truncated to $\sim$ 400 characters). For each submission, we identified if the corresponding problem exists in Codeforces R1-Dataset and we marked if it passes all the public test cases along with the generated tests by DeepSeek-R1.

To preprocess the dataset, we removed all the submissions that satisfied the following conditions:
\begin{itemize}
    \item[]
    \begin{itemize}
     \item All the submissions that were from problems that cannot be found in the Codeforces R1-Dataset.
     \item Buggy submissions that passed all the tests and all the anchors that failed tests.
     \item Buggy submissions that had different C / C++ versions, used for compiling, from the anchor.
     \item Submissions that had verdicts which were different from (Accepted, Runtime Error, Memory Limit Exceeded, Time Limit Exceeded, Wrong Answer and Compilation Error).
     \item Buggy submissions that had a similarity lower than 60\% with the anchor.
     \item All the buggy submissions that did not have an anchor after or before the preprocessing.
    \end{itemize}
\end{itemize}

After preprocessing the dataset, we remain with 567 <anchor, buggy submission> pairs. From the $\sim$ 3000 submissions, $\sim$ 600 submissions didn't have an anchor (no Accepted submissions for that specific problem for a given user), $\sim$ 700 submissions were for problems which could not be found in Codeforces-R1 dataset and $\sim$ 900 submissions were removed due to the similarity threshold. 

\subsection {Metrics}
To be able to quantify how much an LLM is deviating from a buggy submission compared to a human fix, we need to know: (1) the similarity between the buggy solution and the anchor;(2) the similarity between the buggy solution and the generated solution; (3) whether the generated solution passes all the tests.

\subsubsection{Similarity}
Given two source codes $s_1 \in S$ and $s_2 \in S$  with $n$ number of lines, respectively $m$ number of lines, and $S$ being the set of all possible source codes, we are interested in a function $sim:S \times S \rightarrow [0,1]$ that captures the similarity between them. 

In our case, we define $sim(s_1, s_2) = \frac{s_1 \cap s_2}{s_1 \cup s_2}$ as the Jaccard Index between $s_1$ and $s_2$. To compute the Jaccard Index, we use a Diff Tool to determine the number of deleted lines $d$ and added lines $a$ to get from $s_1$ to $s_2$. We can compute the common lines $c$ as $n - d$. In this way $sim(s_1,s_2) = \frac{c}{d + c + a}$. When we compute the similarity, we strip all the lines of white spaces and tabs, remove empty new lines and comments. 

Given a buggy source code $s_b$ and an anchor $s_a$, we define the baseline similarity as $sim(s_b, s_a)$. For a buggy source code and a generated source code, we define the generated similarity as $sim(s_b, s_g)$. 

We define the Similarity Ratio metric as $sim_{ratio} = \frac{sim(s_b, s_g)}{sim(s_b, s_a)}$. The metric captures two scenarios: (1) $sim_{ratio} < 1$: The generated code by the LLM is not as consistent as the human fix (2) $sim_{ratio} > 1$: The generated code by the LLM is more consistent than the human fix. This can happen in situations where a human fix adds more code than necessary. Note that the $sim_{ratio}$ can be greater than 1 if the baseline similarity is low. When we preprocess the dataset we choose 0.6 as threshold to capture cases in which the human fix may be inefficient in terms on modified lines but also to remove cases when a human fix reimplements the whole solution from scratch. Compared to \cite{dai2025less}, which defines the consistency rate as $r / k$, where $r$ indicates the number of lines preserved in the fixed solution and $k$ the total number of lines in the fixed solution, we use a metric based on the Jaccard Index that also penalizes the number of deleted lines from the buggy solution.  

\subsubsection{Pass} To quantify if the LLMs fixes the bug or not, we are interested to see if the generated fix passes all the tests and thus we define the Test pass for a given submission $s$, and $n$ test cases as $TP(s) = \prod_{i=1}^{n} 1_{\hat{t_k}=t_k}$. We choose to use an indicator function instead of a ratio of passing tests because we are interested in seeing if the LLM correctly fixes the bug and solves the problem.

\section{Experimental Setup}
Using the preprocessed dataset, we evaluate 3 strong closed-source LLMs from the OpenAI family (gpt-5-nano, gpt-5-mini, gpt-5.1) using three prompting methods: NaiveBugFix, GenerateScratch, GEPABugFix. For all the methods, we make use of DSPy prompt optimization framework~\cite{khattab2024dspy} to define the prompts. For each LLM, we use a \textit{medium} reasoning effort. In DSPy, we use a Predictor Module instead of ChainOfThought (COT) module, because the models already have an internal reasoning. We limit each LLM to 32k tokens. For gpt-5.1 we also investigate how the model performs with no reasoning. We split the initial preprocessed dataset by problem into 3 folds: (1) train fold - 15 \% problems (2) validation fold - 15 \% problems (3) test fold - 70 \% problems. We chose to split the preprocessed dataset by problem instead of submission to avoid data leakage. We use the train and validation folds for GEPA prompt optimization -- as GEPA is very data efficient and requires a small number of samples, and we test all the models on the test fold.

\subsubsection{NaiveBugFix}
We create a simple prompt that asks the LLM to generate a fix for a buggy submission on a given problem (by generating the entire code), while keeping the number of deleted or added lines as low as possible. The prompt can be seen in Figure \ref{fig:bugfixer-signature}.

\begin{figure}[h]
\begin{signaturebox}
\textbf{NaiveBugFixPrompt} \\[2pt]
\textit{\small Fix the bug in the buggy code for the given competitive programming problem by adding, deleting or modifying as few lines as possible.
In other words, you must adhere to the given buggy code and change it as little as possible to make it work. You can add, delete or modify lines of code, but you cannot rewrite the whole solution. The more lines you change, the more points you lose. You should try to find the bug and fix it, not to rewrite the whole solution.} \\[4pt]
\noindent\rule{\linewidth}{0.4pt} \\[2pt]
{\small
\textbf{Inputs:} \texttt{problem\_description}, \texttt{input\_format}, \texttt{output\_format}, \texttt{examples}, \texttt{note} \textit{(optional)}, \texttt{programmingLanguage}, \texttt{submission\_verdict}, \texttt{buggy\_code} \\[4pt]
\noindent\rule{\linewidth}{0.4pt} \\[2pt]
\textbf{Output:} \texttt{fixed\_code} — The complete fixed code encapsulated in a \texttt{cpp} block.
}
\end{signaturebox}
\caption{The \textsc{NaiveBugFixPrompt}}
\label{fig:bugfixer-signature}
\end{figure}

\subsubsection{GenerateScratch}
We create a simple prompt that asks the LLM to generate a solution for a given problem. Note that in this case, there is no bug fixing, so the $sim_{ratio}$ will be very low. We use this method to identify a lower bound for the $sim_{ratio}$ and to see if there is any performance gain in the pass metric. The prompt can be seen in Figure \ref{fig:generate-scratch-signature}.

\begin{figure}[h]
\begin{signaturebox}
\textbf{GenerateScratchPrompt} \\[2pt]
\textit{\small       Generate a solution for the given competitive programming problem.} \\[4pt]
\noindent\rule{\linewidth}{0.4pt} \\[2pt]
{\small
\textbf{Inputs:} \texttt{problem\_description}, \texttt{input\_format}, \texttt{output\_format}, \texttt{examples}, \texttt{note} \textit{(optional)}, \texttt{programmingLanguage} \\[4pt]
\noindent\rule{\linewidth}{0.4pt} \\[2pt]
\textbf{Output:} \texttt{generated\_code} — The generated code. Please encapsulate it in a \texttt{cpp} block.
}
\end{signaturebox}
\caption{The \textsc{GenerateScratchPrompt}}
\label{fig:generate-scratch-signature}
\end{figure}

\subsubsection{GEPABugFix}
\begin{figure}[h]
    \centering
    \includegraphics[width=0.5\textwidth]{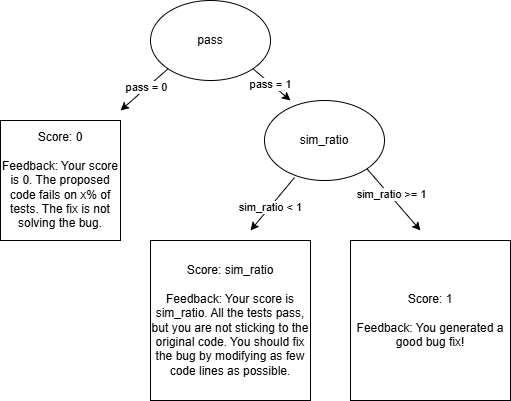} 
    \caption{GEPA Flow}
    \label{fig:GEPAFlow}
\end{figure}

Given the NaiveBugFix as a starting point, we employ GEPA as a prompt optimization technique. We use as a prediction model gpt-5-nano and as a reflection model gpt-5-mini, with medium reasoning, a maximum 32k tokens and 2 full evaluations over the validation dataset. To optimize the prompt, GEPA requires a metric function that returns values between $[0,1]$. To convert our $sim_{ratio}$ metric into a function that has values between 0 and 1, we use the following reward function: $reward = pass * min(1, sim_{ratio})$. Note that $pass \in \{0, 1\}$. Basically, if the generated solution does not fix the bug, then we return 0, since the similarity is not relevant in this case, and thus forcing GEPA to find a prompt that prioritizes bug fixing. If the bug is fixed, then we return the $sim_{ratio}$. Note that if $sim_{ratio} > 1$, we trim it to 1 since we already have a solution that is better or equal with the baseline. GEPA allows for textual feedback in the metric function. The entire flow can be seen in Figure \ref{fig:GEPAFlow}.

\section{Results and Discussion}
For each method and model, we compute the $sim_{ratio}$ of each submission. We group the submissions based on a baseline similarity range to observe how each method / model handles various degrees of similarity. The results can be seen in Table~\ref{table:results}.

\begin{table*}[t]
\centering
\caption{Results for gpt-5-nano, gpt-5-mini, gpt-5.1-no-reasoning, gpt-5.1 for the given methods. The Sim Range column groups the submissions based on the baseline similarity range. $\overline{sim_{ratio}}$ mean is computed based on each submission's $sim_{ratio}$ in the given range. Problems Solved shows the ratio of problems solved. The following acronyms were used: gpt-5-nano (5n), gpt-5-mini(5m), gpt5.1-no-reasoning (5.1nr), gpt5.1 (5.1). For each Similarity Range and each column ($\overline{sim_{ratio}}$, Problems Solved), the bolded text shows the highest score and the underlined text the second highest. }
\label{table:results}
\begin{tabular}{ccccccc|ccccc}
\toprule
\multirow{2}{*}{\textbf{Sim Range}} &  \multirow{2}{*}{\textbf{Samples}} & \multirow{2}{*}{\textbf{Method}}
  & \multicolumn{4}{c|}{\textbf{$\overline{sim_{ratio}}$}}
  & \multicolumn{4}{c}{\textbf{Problems Solved}} \\
\cmidrule(lr){4-7} \cmidrule(lr){8-11}
 & & & \textbf{5n} & \textbf{5m} & \textbf{5.1nr} & \textbf{5.1}
   & \textbf{5n} & \textbf{5m} & \textbf{5.1nr} & \textbf{5.1} \\
\midrule
\multirow{3}{*}{[0.6--0.7)} & \multirow{3}{*}{60}
 & GEPABugFix      & 0.70 & 0.94 & \textbf{1.20} & \underline{1.20} & 0.73 & 0.72 & 0.38 & 0.77 \\
 & & GenerateScratch & 0.14 & 0.13 & 0.13 & 0.15 & \underline{0.87} & \underline{0.87} & 0.63 & \textbf{0.97} \\
 & & NaiveBugFix     & 0.59 & 0.90 & 0.99 & 1.02 & 0.80 & 0.77 & 0.48 & 0.83 \\
\midrule
\multirow{3}{*}{[0.7--0.8)} & \multirow{3}{*}{66}
 & GEPABugFix      & 0.48 & 0.72 & \underline{0.94} & \textbf{0.96} & 0.82 & 0.82 & 0.36 & \underline{0.91} \\
 & & GenerateScratch & 0.11 & 0.11 & 0.12 & 0.12 & 0.89 & \underline{0.91} & 0.64 & \textbf{0.97} \\
 & & NaiveBugFix     & 0.51 & 0.68 & 0.76 & \underline{0.94} & 0.83 & 0.85 & 0.48 & \underline{0.91} \\
\midrule
\multirow{3}{*}{[0.8--0.9)} & \multirow{3}{*}{115}
 & GEPABugFix      & 0.61 & 0.84 & \textbf{0.96} & \underline{0.94} & 0.75 & 0.72 & 0.41 & 0.77 \\
 & & GenerateScratch & 0.10 & 0.10 & 0.10 & 0.11 & \underline{0.92} & \underline{0.92} & 0.71 & \textbf{0.98} \\
 & & NaiveBugFix     & 0.65 & 0.79 & 0.81 & 0.87 & 0.74 & 0.78 & 0.49 & 0.86 \\
\midrule
\multirow{3}{*}{[0.9--1]} & \multirow{3}{*}{115}
 & GEPABugFix      & 0.61 & 0.76 & \underline{0.88} & \textbf{0.93} & 0.75 & 0.83 & 0.58 & 0.90 \\
 & & GenerateScratch & 0.09 & 0.09 & 0.09 & 0.10 & 0.90 & 0.90 & 0.77 & \underline{0.93} \\
 & & NaiveBugFix     & 0.64 & 0.72 & 0.76 & 0.83 & 0.73 & 0.89 & 0.63 & \textbf{0.95} \\
\bottomrule
\end{tabular}
\end{table*}

\subsubsection{RQ1: Given a buggy submission, how well does the LLM preserve the original user's structure when fixing the bug?} For each method and model, when the model capacity is increased (e.g. gpt-5-nano $\rightarrow$ gpt-5-mini), the $\overline{sim_{ratio}}$ increases as well. In submissions with a lower similarity range, GEPABugFix and NaiveBugFix methods seem to find solutions that are even better in preserving the structure than the user fix, while this improvement seem to decrease if the similarity range increases. The high gap between GEPA (0.93) and NaiveBugFix (0.83) on $\overline{sim_{ratio}}$ on the [0.9 - 1] range illustrates that when a simple prompt is used and the bug is very localized, the LLMs modify other unnecessary lines. This suggests that when asked to solve a bug, LLMs tend to alter additional code which is not part of the bug, thus deviating from the original structure. Reasoning does not seem to affect the $\overline{sim_{ratio}}$ in a substantial way, with the exception of the [0.7-0.8) range for NaiveBugFix, which suggests that the LLM is capable of following the instruction to do minimal edits. When generating a problem from scratch, the $\overline{sim_{ratio}}$ is greater than 0 because there is a high probability of having some libraries or the starting point of the program in common.

\subsubsection{RQ2: Are models better at generating code that solves a problem than at fixing bugs?} In the case of gpt-nano, there seems to be $\sim$ 0.15 drop for bug fixing compared to generating from scratch, and for gpt-mini as well in some similarity ranges. For the other models, this drop happens in various similarity ranges as well, with the exception of gpt 5.1 in the [0.9 - 1] range. This suggests that LLMs are better at generating a solution from scratch than at bug fixing. Besides this, we can observe that increasing the model capacity increases the score on a task level but doesn't always reduce the gap between the tasks. This can be observed on the [0.6 - 0.7) and [0.8 - 0.9) ranges. Also, reasoning seems to have a big impact on the ratio of problems solved since gpt-5.1-no-reason has low scores. The superiority of LLMs at generating solutions from scratch over fixing bugs may create a misleading expectation in users. Since debugging intuitively feels like a more constrained and therefore easier task than solving a problem from scratch, users may incorrectly assume that LLMs should excel at it when the evidence suggests the opposite.

\subsubsection{RQ3: Is there a relationship between how much the AI preserves the user's original code structure and its ability to correctly fix the bug?} In the case of GEPABugFix method, even though the $\overline{sim_{ratio}}$ is better than in the case of NaiveBugFix method, the ratio of problems solved is lower. This suggests that if an LLM is forced to fix a bug with minimal modifications, it tends to focus more on the structure instead of actually solving the problem correctly. This could happen because GEPA defines many rules to minimize the number of modified lines that may obscure the task of actually fixing the bug. This forces the user to choose between receiving solutions that have localized patches, which are more likely to be wrong or a correct solution that diverges from the original structure and has to be understood. 

\subsubsection{Discussion summary} - From the results above, it can be seen that the LLMs are better at the task of generating a solution from scratch to a given problem (especially lower capacity models which may be overfitted on the task) than on the task of bug fixing with minimal modifications. Having a complex prompt that forces the model to minimize the number of modified lines when fixing a bug increases the $\overline{sim_{ratio}}$ but decreases the ratio of solved problems while a more simpler prompt decreases the $\overline{sim_{ratio}}$ and increases the number of problems solved. These findings point to open challenges for AI-assisted debugging tools. One promising direction is to adapt how strictly the model is constrained based on how large the fix needs to be — enforcing minimal edits only when the required change is small.

\section{Reproducibility}
The code, GEPA model and dataset can be found on \href{https://github.com/Arkin1/CanLLMFixSubmissions}{Github}. All the experiments were conducted using Docker to ensure reproducibility. Each generated solution was tested against tests by using the solution suggested on the Codeforces-R1 Huggingface page. For the entire study, we have consumed $\sim$ 20 million tokens with a price of $\sim$ 80\$.

\section{Conclusion}
In this paper, we have investigated how LLMs perform when they are required to fix bugs with minimal modifications. We have found that LLMs are better at generating solutions from scratch than at bug fixing with minimal modifications especially when the number of modified lines in the human fix is very small. Besides this, we have also found that  defining a more complex prompt to reduce the number of modifications has a negative impact on the capacity of the model to correctly solve a problem. This suggests that LLMs may not propose  bug fixes that are consistent with the initial code and perform other modifications that the user does not expect. Since LLMs perform better at generating solutions from scratch, they may be tempted to rewrite rather than minimally fix code. This is problematic in any domain where the original structure carries value — such as education, where it supports learning; collaborative development, where it aids code review; or legacy systems, where large rewrites introduce risk and break established conventions.

As future work we would like to extend the study to LLMs from other vendors as well to understand if the effects replicate. We would also like to extend the dataset to contain more users and track how the gap between scratch generation and bug fixing evolves as newer, more recent submissions are introduced. We also want to analyze how the effects manifest based on problem difficulty and judge verdicts.
Besides this, we are also interested to evaluate the model performance on cases in which they are allowed to have more than 1 call per submission and to further explore the impact of prompt complexity on the relation between preserving the structure and fixing the buggy submission.

\bibliographystyle{IEEEtran}
\bibliography{bibfile.bib}

\end{document}